\documentclass{article}

\usepackage[final,nonatbib]{neurips_2020}

\usepackage[utf8]{inputenc} % allow utf-8 input
\usepackage[T1]{fontenc}    % use 8-bit T1 fonts
\usepackage{hyperref}       % hyperlinks
\usepackage{url}            % simple URL typesetting
\usepackage{booktabs}       % professional-quality tables
\usepackage{amsfonts}       % blackboard math symbols
\usepackage{nicefrac}       % compact symbols for 1/2, etc.
\usepackage{microtype}      % microtypography
\usepackage{graphicx}
\usepackage{amsmath}
\usepackage{amsthm}
\usepackage{subcaption}
\usepackage{titlesec}
\usepackage{algorithm}
\usepackage{algorithmic}
\usepackage[inline]{enumitem}

\setlist[enumerate,1]{label=\textbf{(\alph*)}}
\titleformat{\subsubsection}[runin]
{\normalfont}{\thesubsubsection}{1em}{}

\title{Federated Learning with Heterogeneous Labels and Models for Mobile Activity Monitoring}

\author{%
  Gautham Krishna Gudur \\
  Global AI Accelerator, Ericsson \\
  \texttt{gautham.krishna.gudur@ericsson.com} \\
  \And
  Satheesh Kumar Perepu \\
  Ericsson Research \\
  \texttt{perepu.satheesh.kumar@ericsson.com} \\
}

\begin{document}

\maketitle

\sloppy

\begin{abstract}
Various health-care applications such as assisted living, fall detection, etc., require modeling of user behavior through Human Activity Recognition (HAR). Such applications demand characterization of insights from multiple resource-constrained user devices using machine learning techniques for effective personalized activity monitoring. On-device Federated Learning proves to be an effective approach for distributed and collaborative machine learning. However, there are a variety of challenges in addressing statistical (non-IID data) and model heterogeneities across users. In addition, in this paper, we explore a new challenge of interest -- to handle \textit{heterogeneities in labels (activities)} across users during federated learning. To this end, we propose a framework for federated label-based aggregation, which leverages overlapping information gain across activities using \textit{Model Distillation Update}. We also propose that federated transfer of model scores is sufficient rather than model weight transfer from device to server. Empirical evaluation with the Heterogeneity Human Activity Recognition (HHAR) dataset (with four activities for effective elucidation of results) on Raspberry Pi 2 indicates an average deterministic accuracy increase of at least $\sim$11.01\%, thus demonstrating the on-device capabilities of our proposed framework.
\end{abstract}

\section{Introduction}
Human Activity Recognition (HAR) is a technique of significant importance for modeling user behavior in applications like pervasive health monitoring, fitness tracking, fall detection, etc. With the ubiquitous proliferation of personalized sensor-based IoT devices, collaborative and distributed learning is now more possible than ever to help best utilize the behavioral information learnt from multiple users. With the advent of Federated Learning (FL) \cite{cite:federated_learning}, we can effectively train a centralized model with a federation of users without compromising on sensitive data of users by combining local Stochastic Gradient Descent (SGD) of each local (client) model and aggregating weights on a server, instead of conventionally transferring sensitive data to cloud (\textit{Federated Averaging} algorithm) \cite{cite:fedavg}. %Federated learning has been an active area of research in solving problems pertaining to secure communication protocols, optimization, privacy preserving networks, etc. \cite{cite:challenges_methods}.
%Human Activity Recognition (HAR) is an important technique to model user behavior for performing various health-care applications such as health monitoring, fall detection, fitness tracking, etc. Contemporary deep learning has led to major breakthroughs on sensor-based IoT devices due to their automatic feature extraction mechanisms and the compute capabilities vested in resource-constrained mobile and wearable devices. However, such collaborative data sharing might always not be feasible owing to privacy concerns from users who might not prefer sending their private health data to a remote server/cloud.

\begin{figure*}[ht]
  \centering
    \includegraphics[width=195pt, height=160pt]{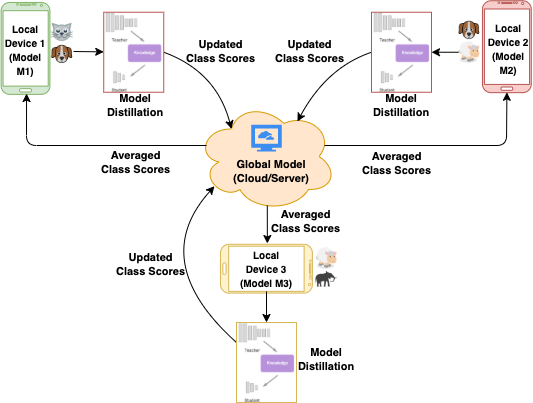}
    \caption{Overall Architecture with \textit{Local Model Distillation}. Each mobile device can consist of disparate set of local labels and models, and they interact with the global model (cloud/server). Each local model is first distilled to a student model, the respective class scores are then aggregated in the global model, and the updated consensus is again distributed across local models.}
    \label{fig:block_diagram}
    \vskip -0.1in
\end{figure*}

On-device federated learning deals with various forms of heterogeneities like device, system, statistical heterogeneities, etc. \cite{cite:challenges_methods}. Particularly, statistical heterogeneities have gained much research visibility predominantly due to the \textit{non-IID} (non-independent and identically distributed) nature of data from distinct distributions, leading to challenges in personalized federation of devices. An important step in this direction is the ability of end-users to have the choice of architecting their own models, rather than using the pre-defined architectures mandated by the global model. FedMD \cite{cite:FedMD} leverages knowledge distillation \cite{cite:knowledge_distillation} to address this, wherein the local models distill their respective knowledge into a \textit{student model} which has a common model architecture. However, as much independence and heterogeneity in architecting the users' own models is ensured in their work, \textit{they do not guarantee heterogeneity and independence in labels across users}.

Many such scenarios with heterogeneous labels and models typically exist in federated IoT settings, such as behaviour/health monitoring, activity tracking, etc. Few works address handling new labels in typical machine learning scenarios, however, to the best of our knowledge, there is no work which addresses this important problem of \textit{label and model heterogeneities} in non-IID federated learning scenarios. More related work on FL, HAR, etc. can be found in Appendix \ref{section:related_work_appendix}.
%This necessitates seamless interaction of the local clients with the global cloud/server independent of model architectures and labels.

\textbf{Scientific contributions:}
\begin{enumerate*}[label=\textbf{(\arabic*)}]
%\begin{itemize}
\item Enabling end-users to characterize their own preferred local architectures in a federated learning scenario for HAR, so that effective transfer learning and federated aggregation happens between global and local models.
%\vspace{0.3cm}
\item A framework to allow flexible heterogeneous selection of activities by showcasing scenarios with and without label overlap across different user devices, thereby leveraging the information learnt across devices pertaining to those overlapped activities.
%\vspace{0.3cm}
\item Empirical demonstration of the framework's ability to handle real-world disparate data/label distributions (non-IID) on simple mobile and wearable devices with a public HAR dataset.
%\end{itemize}
\end{enumerate*}

\section{Our Approach}
\label{section:our_approach}

%In this section, we discuss in detail about the problem formulation of heterogeneity in activities and models, and our proposed framework to handle the same (showcased in Figure \ref{fig:block_diagram}).

%\label{section:problem_formulation}
\textbf{Problem Formulation:} We assume the following scenario in federated learning. There are multiple local devices which can characterize different model architectures based on end users. We hypothesize that the incoming data to different devices also consist of heterogeneities in activities, with either unique or overlapping activities. We also have a public dataset with the label set consisting of all activities -- this can be accessed by any device anytime, and acts as an initial template of the data and labels that can stream through, over different iterations. We re-purpose this public dataset as the test set also, so that consistency is maintained while testing. To make FL iterations independent from the public dataset, we do not expose the public dataset to the local models during learning (training). %The research problem here is to create a unified framework to handle heterogeneous labels and models in a federated learning setting.

\begin{algorithm}[ht]
   \caption{Our Proposed Framework}
   \label{alg:proposed_method}
   \centering
   \begin{algorithmic}
   \STATE \textbf{Input:} Public Dataset $\mathcal{D}_0\{x_0,y_0\}$, Private Datasets $\mathcal{D}_m^i$, Total users $M$, Total iterations $I$, LabelSet $l_m$ for each user\\
   \STATE \textbf{Output:} Trained Model scores $f_G^I$
   %\REPEAT
   \vspace{0.1cm}
   \STATE Initialize $f_G^0 = \mathbf{0}$ (Global Model Scores)
   %\vspace{0.1cm}
   \FOR{$i=1$ \textbf{to} $I$}
   \FOR{$m=1$ \textbf{to} $M$}
   %\vspace{0.1cm}
   \STATE \textbf{Build:} Model $\mathcal{D}_m^i$ and predict $f_{\mathcal{D}_m^i}(x_0)$
   %\vspace{0.1cm}
   \STATE \textbf{Local Update (Model Distillation):}\\
   Build a distilled model only on respective local model labels with global averaged probabilities on public dataset $D_0$. Now, update the model with the new data $\mathcal{D}_m^i$ arriving in this iteration.
   %\vspace{0.1cm}
   \ENDFOR
   %\vspace{0.1cm}
   \STATE{\bfseries Global Update:} Update label wise \\
   $f_G^{i+1} = \displaystyle \sum_{m=1}^{M}\beta_m f_{\mathcal{D}_m^i}(x_0)$, where \\
   $\beta = 
   \begin{cases}
   1 & \text{If labels are unique} \\
   \text{acc}(f_{\mathcal{D}_m^{i+1}}(x_0)) & \text{if labels are not unique}
   \end{cases}$ \\
   where $\text{acc}(f_{\mathcal{D}_m^{i+1}}(x_0))$ is the accuracy function of the given model, and is defined by the ratio of correctly classified samples to the total samples for the given local model
   \ENDFOR
\vskip -0.1in
\end{algorithmic}
\end{algorithm}

%\label{section:proposed_framework}
\textbf{Proposed Framework:} Our proposed framework to handle heterogeneous labels and models in a federated learning setting is presented in Algorithm \ref{alg:proposed_method}. There are three important steps here:

\begin{enumerate}
    \item \textbf{Build}: In this step, we build the model on the incoming data we have in each local user, i.e., local private data for a specific iteration. The users can choose their own model architecture which suits best for the data present in that iteration.
    \item \textbf{Local Update (Model Distillation Update)}: In this step, the local models trained on local private data, and distilled to a student model (with public data) with corresponding labels. Distillation acts like a summarization of information captured from previous FL iterations. The global averaged consensus are distributed to the local models before a local update.
    \item \textbf{\textbf{Global update}}: In this step, \textit{only the scores of each local model are sent to the server} rather than traditional transfer of weights. The model scores are then averaged with parameter $\beta$, where $\beta$ governs the weightage given to overlapping labels across users using test accuracies of corresponding labels on public data. This module gives the global update scores.
\end{enumerate}

\section{Experiments and Results}
\label{section:exp_results}

\begin{table*}[ht]
\vskip -0.1in
\caption{Model Architectures (filters in each layer), Labels and Activity Windows per federated learning iteration across user devices. Note the disparate model architectures and labels across users.}
\label{table:models_labels_iterations}
\centering
\resizebox{13.95cm}{!}{
\begin{tabular}{ccccc}
\hline
\textbf{}                                                                         & \textbf{User\_1}                                                                  & \textbf{User\_2}                                                                   & \textbf{User\_3}                                                               & \textbf{Global\_User}                                                       \\ \hline
\textbf{Architecture}                                                             & \begin{tabular}[c]{@{}c@{}}2-Layer CNN (16, 32)\\ Softmax Activation\end{tabular} & \begin{tabular}[c]{@{}c@{}}3-Layer CNN (16, 16, 32)\\ ReLU Activation\end{tabular} & \begin{tabular}[c]{@{}c@{}}3-Layer ANN (16, 16, 32)\\ ReLU Activation\end{tabular} & –                                                                           \\ \hline
\textbf{Activities}                                                               & \{Sit, Walk\}                                                                     & \{Walk, Stand\}                                                                    & \{Stand, StairsUp\}                                                            & \{Sit, Walk, Stand, StairsUp\}                                              \\ \hline
\textbf{\begin{tabular}[c]{@{}c@{}}Activity Windows\\ per iteration\end{tabular}} & \begin{tabular}[c]{@{}c@{}}\{2000, 2000\}\\ = 4000\end{tabular}                   & \begin{tabular}[c]{@{}c@{}}\{2000, 2000\}\\ = 4000\end{tabular}                    & \begin{tabular}[c]{@{}c@{}}\{2000, 2000\}\\ = 4000\end{tabular}                & \begin{tabular}[c]{@{}c@{}}\{2000, 2000, 2000, 2000\}\\ = 8000\end{tabular} \\ \hline
\end{tabular}
}
\vskip -0.05in
\end{table*}

We simulate a Federated Learning scenario with multiple iterations of incremental data across users' devices. We use the \textit{\textbf{Heterogeneity Human Activity Recognition dataset}} \cite{cite:HHAR} for our experiment. More details regarding the dataset and data preprocessing are discussed in Appendix \ref{section:dataset_appendix}.

\textbf{Label Heterogeneities:} In our experiment, we consider four activities -- \{\textit{Sit, Walk, Stand, StairsUp}\} across three user devices from the dataset (Table \ref{table:models_labels_iterations}). The activities in each local user can either be unique (present only in that single user) or overlapping across users (present in more one user). We split the four activities into three pairs of two activities each for each user, for convenience of showcasing the advantage of overlapping activities. We also create a non-IID environment across different federated learning iterations wherein, the activity data are split with disparities in both the aforementioned labels and distributions in data (\textit{Statistical Heterogeneities}).
%Also, we consider 2000 activity windows per user per iteration.

\textbf{Model Heterogeneities:} We choose three different model architectures for the three different local users (Table \ref{table:models_labels_iterations}). We also use a simple two-layer ANN model with (8, 16) filters as the \textit{distilled student architecture}. To truly showcase near-real-time heterogeneity and model independence, we induce a change in the model architectures across various FL iterations as shown in Appendix \ref{section:changing_models_appendix} Table \ref{table:changing_models}.

We create a Public Dataset ($D_0$) with 8000 activity windows, and 2000 activity windows corresponding to each activity. Also, we consider 2000 activity windows per label per iteration in a user (Table \ref{table:models_labels_iterations}). In total, we run 15 federated learning iterations in this whole experiment, with each iteration running with early stopping (with a maximum 5 epochs). We track the loss using categorical cross-entropy loss function for classification, and use the Adam optimizer \cite{cite:adam_optim} to optimize the classification loss. We simulate all our experiments -- both federated learning and inference on a \textit{Raspberry Pi 2}.

\begin{figure*}[ht]
\centering
    \begin{subfigure}[b]{0.33\textwidth}
        \includegraphics[width=\linewidth]{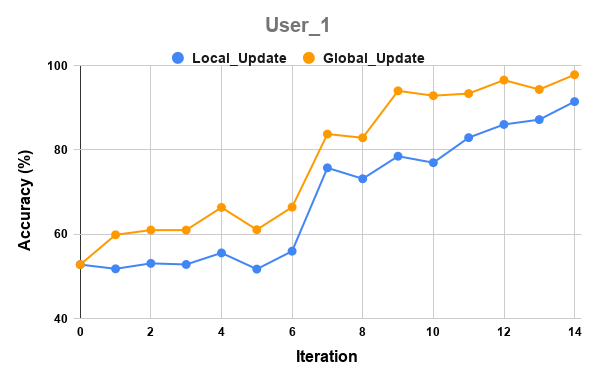}
        \caption{User 1}
        \label{fig:duser1har}
    \end{subfigure}%
    \begin{subfigure}[b]{0.33\textwidth}
        \includegraphics[width=\linewidth]{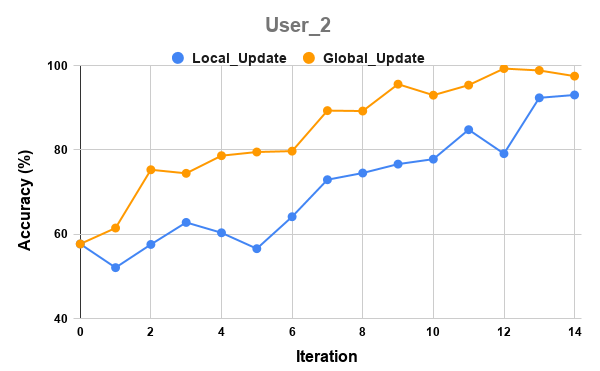}
        \caption{User 2}
        \label{fig:duser2har}
    \end{subfigure}%
    \begin{subfigure}[b]{0.33\textwidth}
        \includegraphics[width=\linewidth]{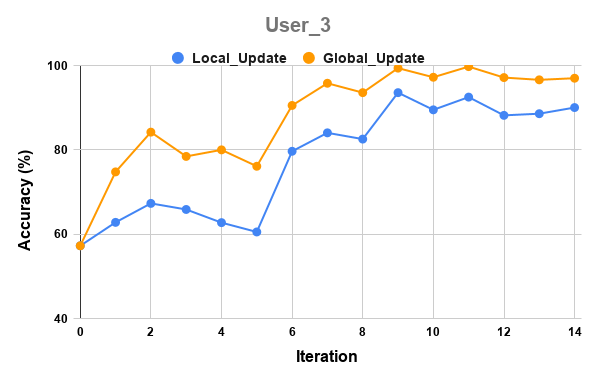}
        \caption{User 3}
        \label{fig:duser3har}
    \end{subfigure}
    \caption{Iterations vs \textit{Local Update} and \textit{Global Update} Accuracies across all three users.}
    \label{fig:hhar_results}
    \vskip -0.1in
\end{figure*}

\subsection{Discussion on Results}
\label{section:discussion_results}

In Table \ref{table:hhar_results}, \textit{Local Update} and \textit{Global Update} signify the accuracies of each local updated model and the corresponding global updated model (after $i^{th}$ iteration) on Public Dataset respectively. We can clearly observe that the global updates (governed by $\beta$), are higher for all three users than the accuracies of their respective local updates. The average accuracy increase across all iterations from local to global updates is deterministically $\sim$11.01\%.

\begin{table}[ht]
\caption{Average Accuracies (\%) of Local and Global Updates, and their respective Accuracy increase.}
\label{table:hhar_results}
\centering
\resizebox{8.5cm}{!}{
\begin{tabular}{llll}
\toprule
& \textbf{Local Update} & \textbf{Global Update} & \textbf{Accuracy Increase} \\ \midrule
\textbf{User\_1} & 68.38                  & 77.61                   & 9.23                      \\
\textbf{User\_2} & 70.82                  & 84.4                    & \textbf{13.58}             \\ 
\textbf{User\_3} & 77.68                  & 87.9                      & 10.22                      \\ 
\textbf{Average} & \textbf{72.293}         & \textbf{83.303}          & \textbf{11.01}              \\
\bottomrule
\end{tabular}}
\vskip -0.05in
\end{table}

The significant contribution of overlap in activities towards increase in global accuracies ($\beta$-weighted global update information gain) is vividly visible in User 2 (Figure \ref{fig:duser2har} -- \{\textit{Walk, Stand}\}), where, in spite of an accuracy dip in local update at iterations 5 and 12, the global updates at those iterations do not spike down. This can be primarily attributed to the robustness of overlapping label information gain from User 1 (\textit{Walk}) and User 3 (\textit{Stand}) on User 2. On the contrary, for User 3 (Figure \ref{fig:duser3har}), when a dip in local update accuracies are observed at iterations 5 and 8, the global update accuracies in those respective iterations also spike down in a similar fashion. Similar trends in local and global accuracies can also be observed in User 1 (Figure \ref{fig:duser1har}). This also accounts for the highest accuracy increase from local to global updates in User 2 over the rest (Table \ref{table:hhar_results}). The final overall global model accuracies averaged across all users after each iteration are also elucidated in Appendix \ref{section:global_accuracy_appendix} Figure \ref{fig:global_results}.

\subsection{On-Device Performance}
\label{section:ondevice}

Raspberry Pi 2 is used for evaluating our proposed framework as it has similar hardware and software (HW/SW) specifications to predominant contemporary IoT/mobile devices. The computation times taken for execution of our proposed framework are reported in Appendix \ref{section:ondevice_appendix} Table \ref{table:timeOnPi}. The distillation mechanism accounts for higher computation time overheads on edge/mobile devices, which depends on the temperature parameter (default set at 1) and the distilled student model architecture chosen.

\section{Conclusion}
\label{section:conclusion}

This paper presents a framework for flexibly handling heterogeneous labels and model architectures in federated learning for Human Activity Recognition. We propose a framework with model distillation in local models, and leverage the effectiveness of global model updates with label based averaging (weighted $\beta$-update) to obtain higher efficiencies. Moreover, overlapping activities across user devices are found to make our framework robust, and also aid in effective accuracy increase. We also experiment by sending only model scores rather than model weights from user device to server, which reduces latency and memory overheads multifold. We empirically showcase the successful feasibility of our framework on-device, for federated learning across different iterations. We expect a good amount of research focus hereon in handling statistical, model and label based heterogeneities for HAR and other pervasive mobile health tasks.

\bibliography{References}
\bibliographystyle{plain}
\pagebreak

\appendix

\section{Appendix: Related Work}
\label{section:related_work_appendix}

Deep learning for HAR, particularly inertial/sensor-based HAR for improving pervasive healthcare has been an active area of research \cite{cite:ploetz_har,cite:Healthcare1}. Particularly, mobile and wearable based deep learning techniques for HAR have proven to be an extremely fruitful area of research with neural network models being able to efficiently run on such resource-constrained devices \cite{cite:DeepSense,cite:iot_wearable,cite:HARNet}. Many other challenges in deep learning for HAR tasks have been explored like handling unlabeled data using semi-supervised mechanisms \cite{cite:sensegan,cite:activeharnet}, domain adaptation \cite{cite:akhil_domainadapt}, few-shot learning \cite{cite:fewshot_har} and many more. 

Federated Learning has contributed vividly in enabling distributed and collective machine learning across various such devices. Federated learning and differentially private machine learning have, or soon will emerge to become the de facto mechanisms for dealing with sensitive data, data protected by Intellectual Property rights, GDPR, etc. \cite{cite:federated_learning}. Federated Learning was first introduced in \cite{cite:fedavg}, and new challenges and open problems to be solved \cite{cite:challenges_methods} and multiple advancements \cite{cite:advances} have been proposed and addressed in many interesting recent works.

Multiple device and system heterogeneities making them optimization problems are addressed \cite{cite:fedprox_heterogeneity}. Personalized federated learning closely deals with optimizing the degree of personalization and contribution from various clients, thereby enabling effective aggregation as discussed in \cite{cite:personalized_fed}. Federated learning on the edge with disparate data distributions -- non-IID data, and creating a small subset of data globally shared between all devices is discussed in \cite{cite:federated_non_iid}.

Particularly for Federated Learning in IoT and pervasive (mobile/wearable/edge) devices, important problems and research directions on mobile and edge networks are addressed in this survey \cite{cite:federated_mobile_survey}, while federated optimization for on-device applications is discussed in \cite{cite:federated_opt}. Federated Learning for HAR is addressed in \cite{cite:fed_har1} which deals with activity sensing with a smart service adapter, while \cite{cite:fed_har_2} compares between centralized and federated learning approaches.

FedMD \cite{cite:FedMD}, which we believe to be our most closest work, deals with heterogeneities in model architectures, and addresses this problem using transfer learning and knowledge distillation \cite{cite:knowledge_distillation}, and also uses an initial public dataset across all labels (which can be accessed by any device during federated learning). Current federated learning approaches predominantly handle same labels across all users and do not provide the flexibility to handle unique labels. However, in many practical applications, having unique labels or overlapping labels for each local client/model is a very viable scenario owing to their dependencies and constraints on specific regions, demographics, privacy constrains, etc. A version of the proposed work is discussed for vision tasks in \cite{cite:aiot20} and for HAR tasks in \cite{cite:dlhar_ijcai20}.
%To the best of our knowledge, none of the works take into account, label and model heterogeneities, particularly in the context of HAR.

\section{Appendix: Dataset and Dataset Preprocessing}
\label{section:dataset_appendix}

The Heterogeneity Human Activity Recognition dataset \cite{cite:HHAR} consists of inertial data from four different mobile phones across nine users performing six daily activities: Biking, Sitting, Standing, Walking, Stairs-Up, Stairs-Down in heterogeneous conditions.

\textbf{Data Preprocessing:} In this experiment, we perform similar preprocessing techniques as stated in \cite{cite:HARNet}. As discussed in \cite{cite:HARNet}, we use the mobile phone accelerometer data only and not gyroscope, due to the reduction in data size without substantial accuracy decrease. We initially segment the triaxial accelerometer data into two-second non-overlapping windows and then perform \textit{Decimation} to downsample (normalize) all activity windows to the least sampling frequency (50 Hz). Following this, \textit{Discrete Wavelet Transform (DWT)} is performed for obtaining temporal and frequency information and we use Approximation coefficients only, all together is stated to have a substantial decrease in data size with much loss in information.

\section{Appendix: Model Heterogeneities across Iterations}
\label{section:changing_models_appendix}

The model heterogeneities across and within different iterations are observed in Table \ref{table:changing_models}.

\begin{table}[ht]
\caption{Details of Model Architectures (filters in each layer) changed across federated learning iterations and users for both datasets.}
\label{table:changing_models}
\centering
\begin{tabular}{ll}
\toprule
\textbf{Iteration}    & \textbf{New Model Architecture}                      \\
\midrule
User\_1 Iteration\_10 & \begin{tabular}[c]{@{}c@{}}3-Layer ANN\\ (16, 16, 32)\\ ReLU Activation\end{tabular}            \\ \hline
User\_1 Iteration\_14 & \begin{tabular}[c]{@{}c@{}}1-Layer CNN\\ (16)\\ Softmax Activation\end{tabular}                 \\ \hline
User\_2 Iteration\_6  & \begin{tabular}[c]{@{}c@{}}3-Layer CNN\\ (16, 16, 32)\\ Softmax activation\end{tabular}         \\ \hline
User\_3 Iteration\_5  & \begin{tabular}[c]{@{}c@{}}4-Layer CNN\\ (8, 16, 16, 32)\\ Softmax activation\end{tabular}   \\
\bottomrule
\end{tabular}
\end{table}

\section{Appendix: On-Device Performance}
\label{section:ondevice_appendix}

The computation times for on-device performance on Raspberry Pi 2 for our proposed framework are observed in Table \ref{table:timeOnPi}.

\begin{table}[ht]
\caption{Computation Time taken for Execution for HHAR dataset}
\label{table:timeOnPi}
\centering
\begin{tabular}{lll}
\toprule
\textbf{Process}    & \textbf{Computation Time}   \\
\midrule
\begin{tabular}[c]{@{}c@{}}Training time per epoch\\ in an FL iteration ($i$)\end{tabular}  & $\sim$1.7 sec   \\
\hline
Inference time                                                                              & $\sim$15 ms        \\
\hline
Discrete Wavelet Transform                                                                  & $\sim$0.45 ms      \\
\hline
Decimation                                                                                  & $\sim$4.6 ms       \\
%\hline
%Knowledge Distillation                                                                      & $\sim$13 sec       \\
\bottomrule
\end{tabular}
\end{table}

\section{Appendix: Overall Global Average Accuracies}
\label{section:global_accuracy_appendix}

\begin{figure*}[ht]
%\vskip 0.1in
\centering
    \includegraphics[width=0.7\linewidth, height=190pt]{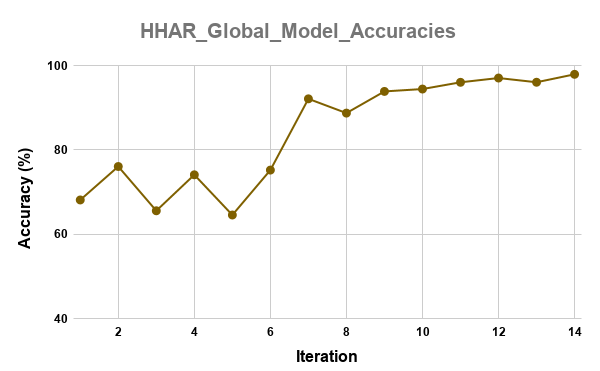}
    \caption{Iterations vs Final Global Average Accuracies (\%) with \textit{Local Model Distillation}.}
    \label{fig:global_results}
%\vskip -0.1in
\end{figure*}

The final overall global model accuracies averaged across all users in each iteration is observed in Figure \ref{fig:global_results}.

\end{document}